\documentclass[letterpaper, 10 pt, conference]{ieeeconf}  % Comment this line out if you need a4paper

\IEEEoverridecommandlockouts                              % This command is only needed if 
\usepackage{graphics} % for pdf, bitmapped graphics files
\usepackage{graphicx}
\usepackage{etoolbox}
\usepackage{epsfig} % for postscript graphics files

\title{\LARGE \bf
Analyzing Reluctance to Ask for Help When Cooperating With Robots: Insights to Integrate Artificial Agents in HRC
}

\author{Ane San Martin$^{*}$, Michael Hagenow, Julie Shah, Johan Kildal, and Elena Lazkano% <-this % stops a space
\thanks{*Corresponding author}% <-this % stops a space
\thanks{A.S.M, M.H and J.S are with the Interactive Robotics Group,
        Massachusetts Institute of Technology, belonging to the Department of Aeronautics and Astronautics,
        Cambridge, Massachusetts, USA
        ({\tt\small [anesm03|hagenow]@mit.edu, julie\_a\_shah@csail.mit.edu})}%
\thanks{A.S.M and J.K are with the Department of Autonomous and Intelligent Systems,
        Tekniker,
        Eibar, Gipuzkoa, Spain
        ({\tt\small [ane.sanmartin|johan.kildal]@tekniker.es})}
\thanks{E.L is with the Faculty of Informatics,
        UPV/EHU,
        San Sebastian, Gipuzkoa, Spain
        ({\tt\small e.lazkano@ehu.eus})}}%

\begin{document}

\maketitle
\thispagestyle{empty}
\pagestyle{empty}

%%%%%%%%%%%%%%%%%%%%%%%%%%%%%%%%%%%%%%%%%%%%%%%%%%%%%%%%%%%%%%%%%%%%%%%%%%%%%%%%
\begin{abstract}

As robot technology advances, collaboration between humans and robots will become more prevalent in industrial tasks. When humans run into issues in such scenarios, a likely future involves relying on artificial agents or robots for aid. This study identifies key aspects for the design of future user-assisting agents. We analyze quantitative and qualitative data from a user study examining the impact of on-demand assistance received from a remote human in a human-robot collaboration (HRC) assembly task. We study scenarios in which users require help and we assess their experiences in requesting and receiving assistance. Additionally, we investigate participants’ perceptions of future non-human assisting agents and whether assistance should be on-demand or unsolicited. Through a user study, we analyze the impact that such design decisions (human or artificial assistant, on-demand or unsolicited help) can have on elicited emotional responses, productivity, and preferences of humans engaged in HRC tasks.

\end{abstract}

%%%%%%%%%%%%%%%%%%%%%%%%%%%%%%%%%%%%%%%%%%%%%%%%%%%%%%%%%%%%%%%%%%%%%%%%%%%%%%%%

\section{INTRODUCTION}
The increased availability of robot teammates (e.g., collaborative robots) will create work settings without human teammates, where assistance comes from artificial agents \cite{gibney2024ai,jeon2024can}. While this shift can offer benefits like increased efficiency and safety, it also raises concerns. A lack of timely assistance can lead to task stagnation, increasing cognitive load and stress, which harm productivity and mental health \cite{angelidis2019m}. Prolonged stress may even contribute to conditions like anxiety, depression, and gastrointestinal illnesses \cite{van2005can, duan2021association}.

However, even if timely intelligent assistance is available in future human-robot teaming scenarios, will humans use it? Previous work suggests that people often struggle to ask for help, since they usually rely on other workers with greater expertise \cite{thomas2017students}. Requesting help, especially in hierarchical work environments, can be challenging, as individuals may avoid seeking assistance from superiors to prevent appearing incompetent \cite{milliken2003exploratory, singh2000performance, landy1994work}, which not only impacts performance but also increases stress and inefficiency. This reluctance is even more pronounced in human-robot collaboration (HRC) tasks, where the absence of collocated human assistance can lead to feelings of isolation and increased cognitive load, delaying problem-solving and reducing productivity \cite{ferreira2021decision,golden2008impact}. The perceived inconvenience or fear of judgment further degrades performance and heightens error likelihood \cite{bohns2010didn,lee1997going}. Additionally, the cognitive load of managing complex tasks while seeking help can overwhelm a person's working memory, reducing efficiency and increasing stress \cite{chen2023cognitive,kunasegaran2023understanding,boksem2008mental}.

Integration of AI assistance into these scenarios offers potential solutions to these issues. Compared to human assistance, AI and robots can provide non-judgmental and on-demand support, which can reduce the stress and stigma associated with seeking help from humans \cite{kavitha2024beyond}. However, researchers have not analyzed how users behave when they need help or face challenges in HRC scenarios, which is essential for designing robots that truly improve existing issues, such as unpredictability and continuously changing environment.

In our research, we investigate perceptions toward intelligent assistance in human-robot collaboration tasks, by analyzing dynamics related to assistance in a representative HRC assembly task. To explore user attitudes toward potential assistance from artificial agents, we conducted a preliminary study with remote human assistance to understand how seeking help influences user performance and satisfaction. Unlike previous studies that focused on human alone or human-human interactions  \cite{ferreira2021decision, bohns2010didn}, or 
directly deployed robots for assistance \cite{baraglia2017efficient,patel2022proactive}, we aim to analyze the challenges users face nowadays in HRC scenarios when they get stuck and need to ask for help. By examining these aspects, we seek to identify key factors that can help in the design of future AI systems. Our study involved a collaborative assembly task where 20 users constructed a 3D puzzle in collaboration with a robot and with access to on-demand human assistance. To ensure that participants requested help, the task was designed with intentional errors and ambiguities. We analyzed users' perceptions of on-demand human assistance during this human-robot teaming task and the implications of assistance on performance and user experience. Through interviews, we explored users' attitudes towards being assisted by a robot. In this paper, we discuss the study’s implications for future systems and propose approaches for artificial assistance methods.

\section{RELATED WORK}
The absence of collocated human assistance can lead to feelings of isolation and increased cognitive load, which negatively affect problem-solving efficiency and task productivity \cite{ferreira2021decision, golden2008impact}, leading to stressful scenarios. Chronic stress from such scenarios has been linked to adverse health outcomes, including cardiovascular diseases \cite{rishab2022stress}. Delays in help exacerbate frustration and anxiety, further impacting well-being and performance \cite{adler2015effects, golden2008impact}.

These challenges are particularly pronounced in HRC tasks due to their dynamic and unpredictable nature, which often require real-time decision-making, adaptation to changing environments, and communication needs between human and robot partners \cite{goodrich2008human, hoffman2004collaboration}. Unlike traditional automation, HRC tasks demand flexible, context-aware assistance capable of adapting to human behavior and preferences, ensuring effective coordination where both parties understand and respond to each other's actions and intentions.

These challenges have motivated some early research on AI-based robotic assistance, though only a few methods have been previously proposed in the literature. Robots are appreciated for their efficiency, reliability, and capacity to provide consistent support in collaborative environments \cite{chi2021artificially, cha2016nonverbal}. Recent work highlights the promise of proactive assistance, where robots not only respond to user commands but also anticipate user needs to deliver timely, context-aware support. This assistance can range from reminding users of upcoming tasks or goals to suggesting corrective actions upon detecting errors, adjusting task complexity, and even providing motivational or emotional support. For instance, Baraglia \textit{et al.} \cite{baraglia2017efficient} showed how robots recognizing user intentions can proactively suggest the correct tool or step, leading to more efficient task performance. Patel \textit{et al.} \cite{patel2022proactive, patel2023predicting} analyzed how predictive modeling of user routines enabled timely reminders and adjustments, reducing cognitive load.

However, the potential of robots to interpret human emotions and offer empathetic responses remains poorly researched. User acceptance of robotic assistance is influenced by factors such as perceived reliability, ease of use, and the robot’s apparent intelligence \cite{venkatesh2003user, bitkina2020perceived}. Yet, with few exceptions \cite{seo2021emergence}, these factors remain largely unexplored in relation to unpredictable, real-world HRC contexts. Although proactive AI systems offer task-specific support, they often fall short when flexible, context-aware interactions are needed. Fong \textit{et al.}’s survey \cite{fong2003survey} shows that trust, perceived risk, and satisfaction remain critical factors, underscoring the ongoing need for human presence in such scenarios. As a result, in the absence of AI systems able to offer nuanced, empathetic, and adaptive support, human presence is still the default option for providing assistance. 

To design effective proactive systems, it helps to first examine how humans naturally give and receive assistance in collaborative settings. Human-to-human interaction offers a gold standard of support behaviors, such as clarifying instructions, providing real-time feedback, troubleshooting errors, and offering strategic guidance, while also revealing the social and emotional dimensions of help-seeking. By observing these patterns, designers can identify the specific requests and cues that an AI system should be equipped to handle human needs. In other words, understanding how humans assist humans in HRC tasks lays the groundwork for AI, which will have to anticipate similar needs and respond appropriately, ultimately boosting both the efficiency and satisfaction of HRC tasks.

\section{USER STUDY}

Our study aimed to explore user perceptions and behaviors when seeking assistance in HRC tasks from someone not physically present. Through this user study, we examined the challenges of such scenarios and attitudes toward AI robotic assistance, simulating a typical HRC setup where both parties collaborate on an assembly task without on-site supervision.

%%%%%%%%%%%%%%%%

\subsection{Participants}
We recruited 20 participants (10 male and 10 female) with age ranging from 22 to 52 ($M=28, SD= 6.26$), who volunteered to take part in the study. The study was approved by the university institutional review board (IRB2402001209). Thirteen participants had some previous experience working with robots: 4 had collaborated with robots, 8 had programmed them, and 1 had both collaboration and programming experience.
\begin{figure}[b]
\centering
      \includegraphics[scale=0.36]{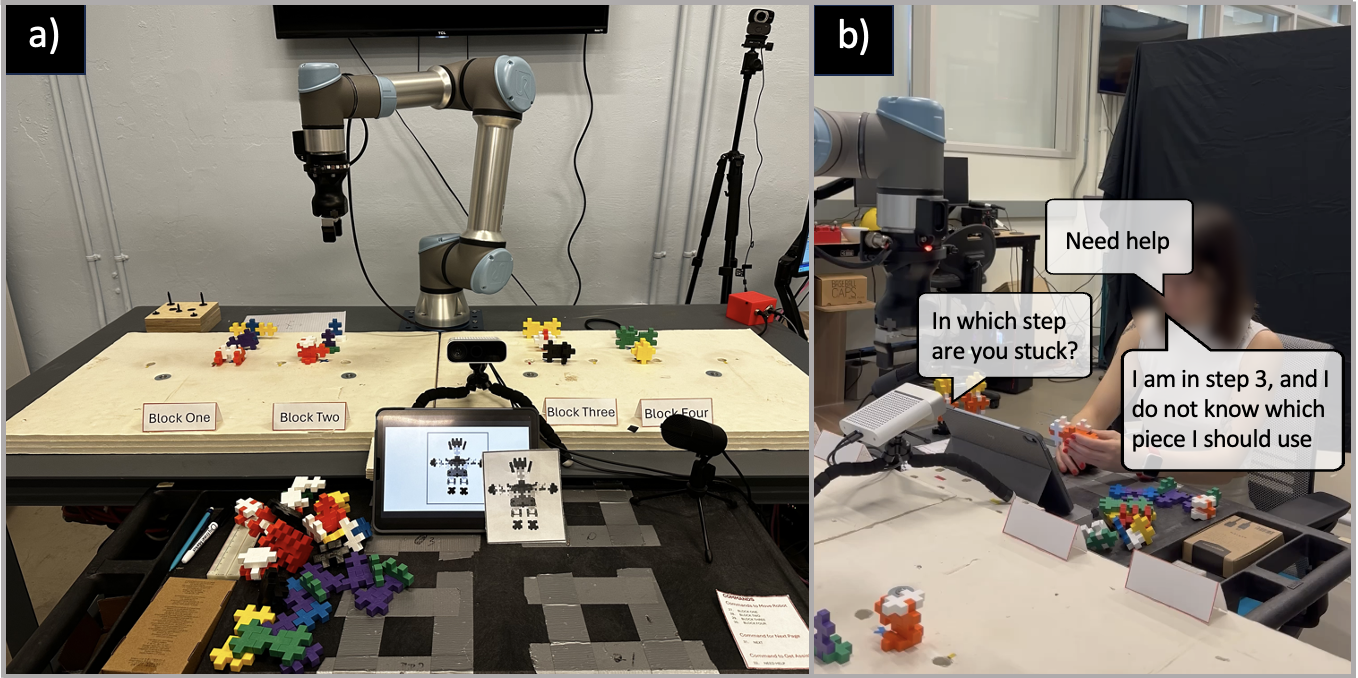}
      \caption{This image shows the experimental setup used in the user study. (a) The robot-assisted assembly task and components involved. (b) A representative interaction where a user receives interactive assistance after encountering difficulties.}
      \label{fig1}
\end{figure}

%%%%%%%%%%

\subsection{Experimental Task}
We designed a single-condition study where the user worked alone alongside a robot. The task instructions were intentionally ambiguous to ensure participants would need assistance at least once during interactions with the robot.

\begin{figure}[tb]
\centering
      \includegraphics[scale=0.4]{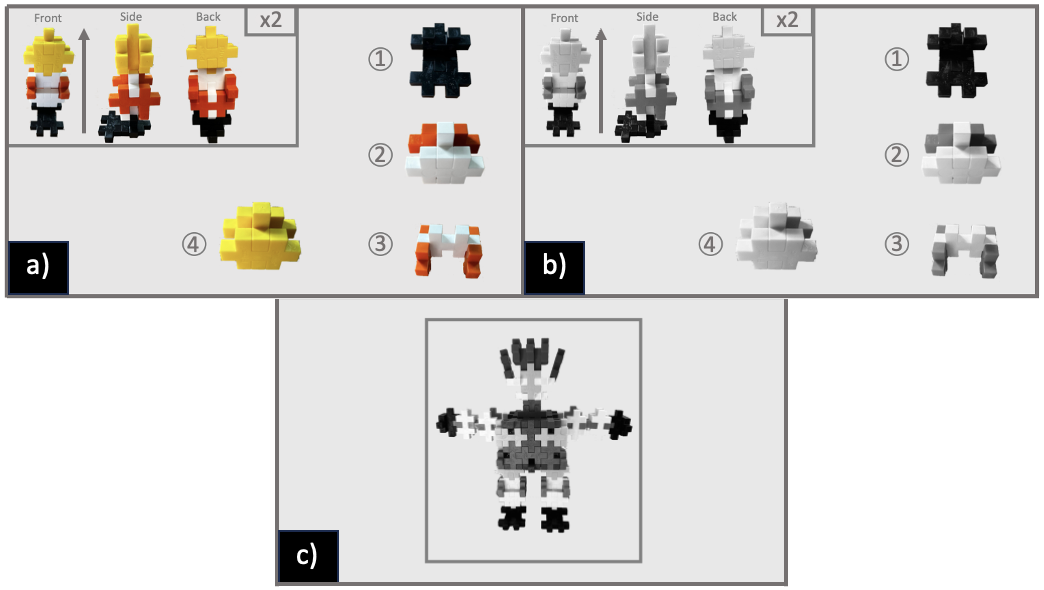}
      \caption{Example of subassembly instructions: (a) Original colored instructions, (b) grayscale instructions presented to users. Numbers on each piece indicated the assembly order, supported by arrows showing construction direction. "x2" denoted structures to be assembled twice. (c) Resulting assembly figure.}
      \label{fig2}
\end{figure}

The human-robot collaboration task involved building a 3D robot puzzle (Fig.~\ref{fig2} (c)) by following assembly instructions on a tablet, mirroring typical industrial HRC assembly processes \cite{Caiazzo2024Comparative, radi2024empowering}. For this scenario, we used the widely adopted UR5 collaborative robot \cite{buerkle2023histodepth}.

To ensure that users needed help at least once, a piece was hidden in a cardboard box on the table (Fig.~\ref{fig1}, bottom left side). The task required assembling 32 pieces and the 23rd piece was hidden. We chose this location to avoid conditioning users early on and to minimize help-seeking at the end of the task, where participants might be fatigued. Grayscale assembly instructions were used (Fig.~\ref{fig2} (b)) to further increase the likelihood of help requests by removing color cues.

\begin{figure}[tb]
\centering
\includegraphics[width=0.5\textwidth]{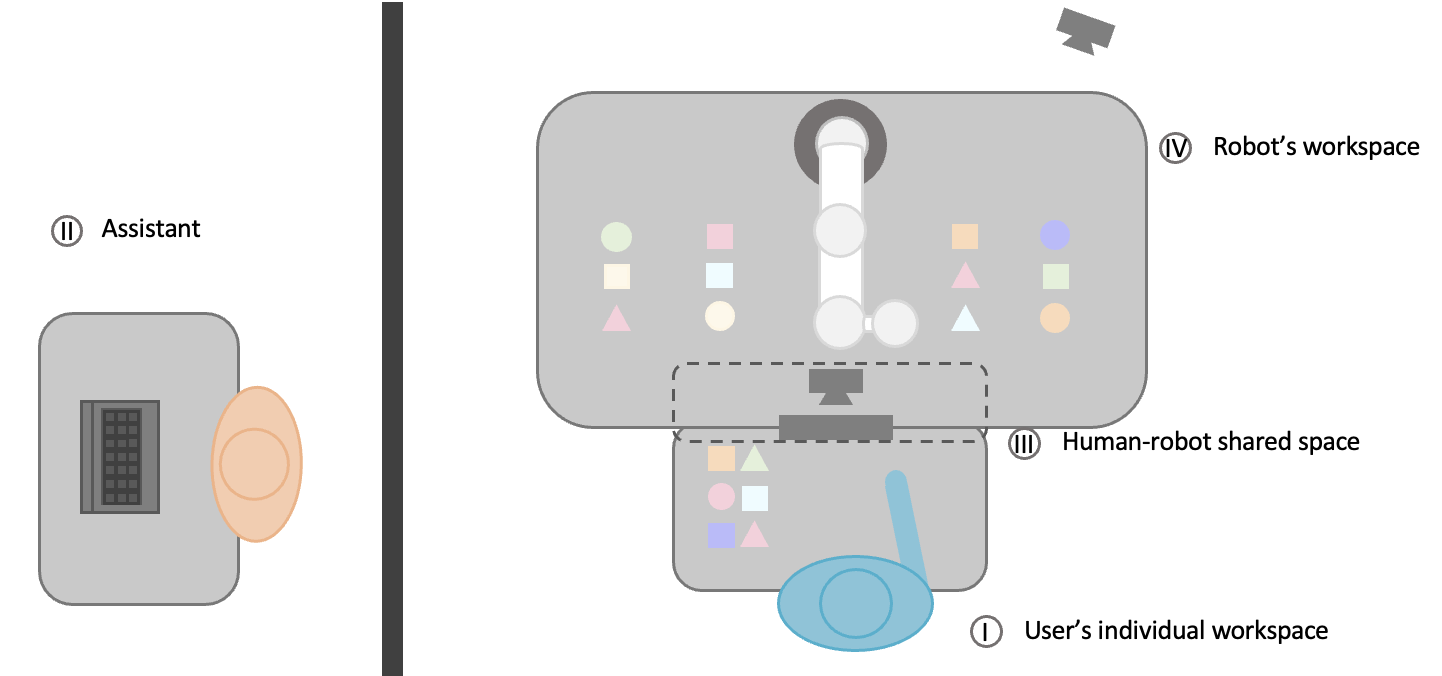}
\caption{The schematic shows the study setup, with participants working in front of the robot, checking if the needed piece was in their area (I) or the robot's area (IV). For pieces in the robot's area, users commanded the robot to deliver them to the designated area (III). Cameras recorded the environment for post-study analysis, which was not available to the assistant. Participants requested help by saying "NEED HELP," notifying the assistant outside the room (II) for online support.}
\label{fig3}
\end{figure}

To perform the task, participants sat at a desk in front of the robot (see Fig.~\ref{fig3} (I)) and used a tablet to view the subassembly tasks (see the example in Fig.~\ref{fig2} (b) and tablet in Fig.~\ref{fig1}). Instructions of each subassembly task provided information about the required pieces and the assembly order, indicated by numbered pieces and arrows that indicated the order of construction (Fig.~\ref{fig2} (b)), although users were allowed to choose their own strategies.

For each step, participants determined whether the required piece was in their or the robot's workspace. Each piece was composed of pre-assembled blocks (Fig.~\ref{fig2}) that participants were instructed to use as given. If the piece was on the robot's side, participants identified its location and said one of four voice commands: \textit{Block One}, \textit{Block Two}, \textit{Block Three}, or \textit{Block Four} (see written commands in Fig.~\ref{fig1} (a)). The robot's pieces were arranged so that users only needed the closest piece in each column and thus could command the robot to fetch the desired piece by identifying the column (Fig.~\ref{fig3} (III)). Participants could choose to collect all needed pieces first or create the assembly step by step.

After completing a task, participants said "NEXT" and the tablet displayed the next subassembly task, continuing until the 3D robot was fully assembled. After saying "NEXT", users could not return to previous steps. If at any point users were unsure how to proceed, they were instructed to say "NEED HELP", prompting verbal help from the external assistant. The assistant, with no access to camera recordings or real-time visuals, relied solely on the pieces requested by users and the current subassembly task of the user. This design choice was made to remove the pressure that might arise from users feeling constantly observed. The assistant would first ask users which specific piece they were struggling with, in current subassembly task and then provide the corresponding assistance.

The experimental setup used a Windows computer, which ran \textit{psi}\footnote{https://www.microsoft.com/en-us/research/project/platform-situated-intelligence/} for camera recording and voice command recognition, with Ubuntu via WSL2 to send robot position commands through ROS. A second computer, equipped with Ubuntu, and running ROS, controlled the robot. ROS served as the communication framework between both computers.

%%%%%%%%%%%%%%

\subsection{Experimental procedure}
Upon arrival, participants were given the informed consent form to sign and a detailed description of the task, along with the opportunity to ask any questions. Then they completed a demographic questionnaire, followed by a training session. During the training session, they practiced voice commands and familiarized themselves with the task dynamics on a different practice assembly task, to avoid learning effects. The training was designed to last 10 minutes.

Before starting the study, participants could ask additional questions. They were also reminded of the available commands, that the assistant could not view camera feeds, that pieces could not be mixed or break-down, and that assistance was only available upon saying "NEED HELP". The main task took users between 20 and 35 minutes to complete. Upon completion the task, the facilitator returned to the room and conducted a semi-structured interview. Once finished, participants received a \$20 Amazon gift card.

%%%%%%%%%%%%

\subsection{Data Collection and Analysis} \label{sec3.4}
For this study, we employed a mixed-method approach, combining qualitative and quantitative measurements. To assess user assistance levels, the video footage from both cameras was categorized into four levels, reflecting varying degrees of assistance required over the duration of the task. Labeling was conducted in \textit{psiStudio}\footnote{https://github.com/microsoft/psi/wiki/Psi-Studio} by two investigators. The first level, the \textit{flow} state, as described by Csikszentmihalyi \cite{czikszentmihalyi1990flow}, represents full immersion and engagement in the task. Levels 1 through 3 indicate increasing assistance needs, as previously defined by Rogers \cite{rogers1989performance}. For coding purposes, we applied the following assistance level definitions:
\begin{itemize}
    \item Flow: The participant is engaged in the task, with no signs of needing assistance.
    \item Level 1: The participant exhibits social cues like looking around or pausing, looking back and forth, indicating they might need help.
    \item Level 2: The participant recognizes the break in Flow and tries to resolve it using external resources, often switching their gaze between the structure and images.
    \item Level 3: Similar to level 2, but the participant finds other sources insufficient and verbally seeks help.
\end{itemize}

After each experiment, a semi-structured interview was conducted to gather qualitative data, focusing on: \textit{how users felt when asking for help}, \textit{challenges they faced when assisted by a human}, and \textit{attitudes toward robotic assistance}. All interviews were first recorded and then transcribed for further analysis. We then thematically organized user comments using Post-it notes, following Lucero's methodology \cite{lucero2015using}.

The execution of the tasks and the post-task interview was performed by a single study coordinator. However, the data analysis, labeling, and the creation of the affinity diagram involved an additional investigator.

%%%%%%%%%%%%%
\section{RESULTS}\label{sec4.0}

%%%%%%%%%%%%%%%
\subsection{Quantitative Results}\label{sec4.1}
Two of the 20 participants were unable to complete the task, and another two finished it without the missing piece, as they did not request help when they noticed it was missing. However, data from all participants were included and analyzed, as their help-seeking behaviors provided valuable insights for the analysis. Users spent a mean time of 26.3 minutes ($SD = 4.4$) $CI_{95\%}$[24.4, 28.2] to complete the task.

\begin{table}[t]
\caption{Numerical values for means, standard deviations (in parentheses), and 95\% confidence intervals for the quantitative data on different levels of needing help.}
\begin{tabular}{l||c|c|c|c}
                  & \textbf{Level 1} & \textbf{Level 2} & \textbf{Level 3} & \textbf{Total}  \\ \hline
                \textbf{\begin{tabular}[l]{@{}l@{}}Time needing\\ help (min)\end{tabular}}        & \begin{tabular}[c]{@{}c@{}}5.2 (4.4)\\ {[}4.4, 6.1{]}\end{tabular} & \begin{tabular}[c]{@{}c@{}}1.3 (0.9)\\  {[}0.7, 1.9{]}\end{tabular}  & \begin{tabular}[c]{@{}c@{}}0.9 (4.8)\\  {[}0.3, 1.6{]}\end{tabular} & \begin{tabular}[c]{@{}c@{}}7.5 (2.5) \\ {[}4.4, 10.2{]}\end{tabular}  \\ \hline
                \textbf{\begin{tabular}[l]{@{}l@{}}\% of the task\\ in each level\end{tabular}}       & \begin{tabular}[c]{@{}c@{}}19.5 (0.1) \\ {[}19.5, 19.6{]}\end{tabular} & \begin{tabular}[c]{@{}c@{}}4.9 (3.1)\\ {[}2.9, 7.0{]}\end{tabular}  & \begin{tabular}[c]{@{}c@{}}3.6 (2.7)\\ {[}1.2, 6.0{]}\end{tabular} & \begin{tabular}[c]{@{}c@{}}28.1 (7.2)\\ {[}20.1, 36.0{]}\end{tabular}    \\ \hline
\end{tabular}
\label{tab1}
\end{table}

As explained, levels 1–3 indicate instances where participants needed help. We calculated the time spent needing help in each level and in total, as well as the percentages (see Table~\ref{tab1}).

Participants spent most time in level 1, frequently switching between task instructions and the paused assembly. Level 2 accounted for the second-most time, characterized by a noticeable pause in task progress without a help request. Level 3 was the least frequent. Two participants did not request help, resulting in 0 minutes at this level. All this resulted in an average of 28.06\% of the task time where users were showing signals of needing help.

Regarding the time showing help signals on a task step, users spent a mean of 34.72 s ($SD = 39.27$) $CI_{95\%}$[29.18, 40.26]. Notably, 17 users were stuck at some point (i.e., showing signs that they were needing help) for around one minute, and 9 of them for more than 2 minutes. Regarding the mean time from realizing a need for help to requesting it, it took a mean of 2.1 minutes ($SD = 1.12$) $CI_{95\%}$[1.68, 2.52], with four users taking over 3 minutes to ask for help. This aspect accounted for a mean of 12.2\% ($SD = 0.01$) $CI_{95\%}$[8.7, 15.6] of the total task time. Observing users' behavior, two participants never requested help, ten asked once, seven asked twice, and one asked three times.

It is noteworthy that 70\% of users displayed clear frustration and stress signs during levels 2 and 3 of help-seeking, with behaviors such as altered facial expressions, face-touching, scanning the environment, and even panting. Two of them even stared at the robot looking frustrated. Four participants, unable to complete the task, continued working despite their frustration, resulting in two outcomes: (1) constructing something entirely different to the requested model by breaking and reassembling pieces or (2) completing the robot but leaving one piece missing.
%%%%%%%%%%%%%%%%%%%%%%%%%

\subsection{Qualitative Results} \label{sec4.2}
\begin{figure*}[tp]
\centering
\includegraphics[width=0.8\textwidth]{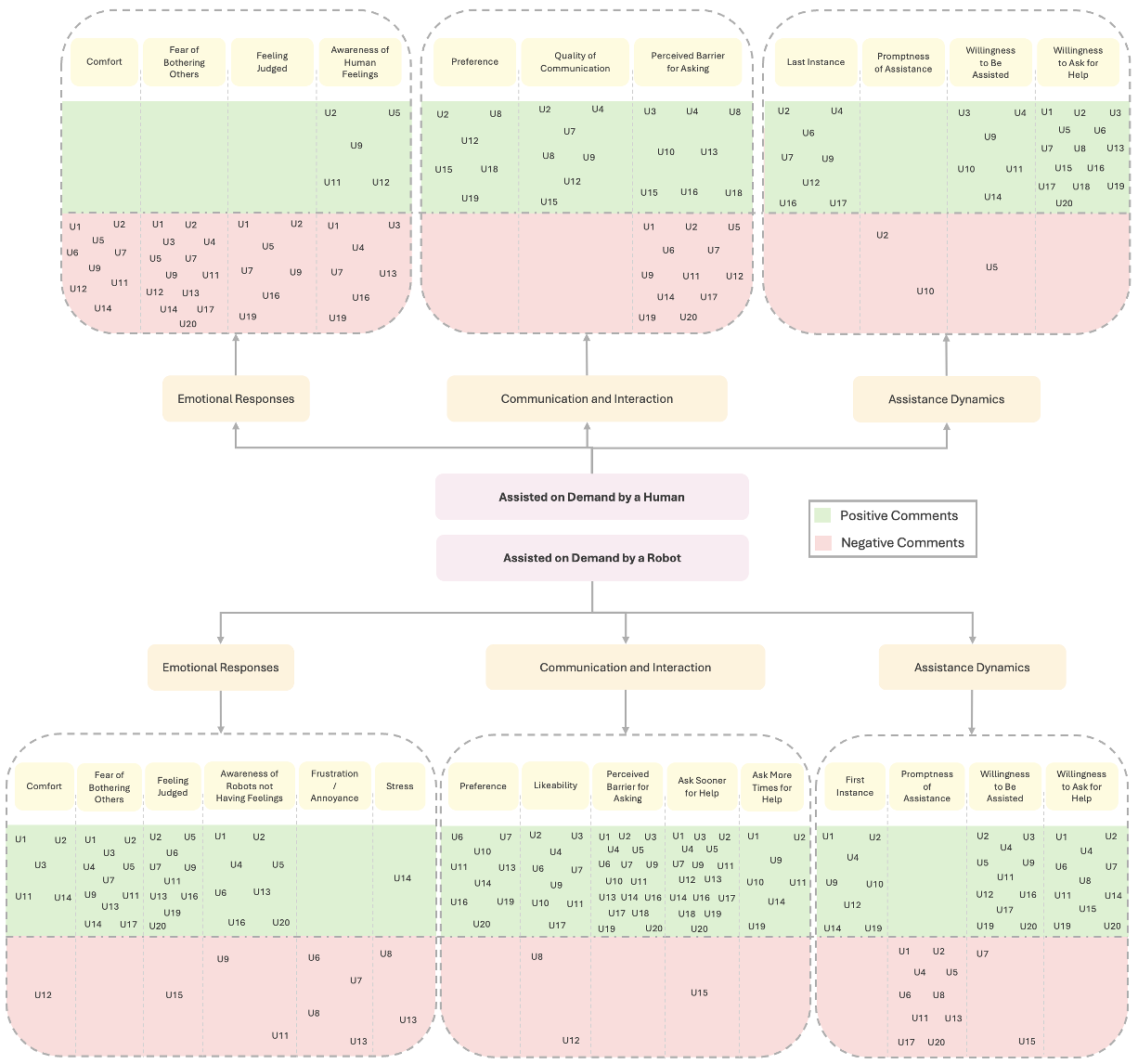}
\caption{Graph obtained from the affinity diagram, based on participant statements from semi-structured interviews. It analyzes on-demand assistance by humans and robots. It is divided into three groups: (1) \textit{Emotional Response} (feelings about being assisted), (2) \textit{Communication and Interaction} (preferences, communication style, and issues), and (3) \textit{Assistance Dynamics} (help-seeking and assistance preferences). The graphs show the number of positive (green) and negative (red) comments for each category in the study.}
\label{fig5}
\end{figure*}

\begin{figure*}[tp]
\centering
\includegraphics[width=0.7\textwidth]{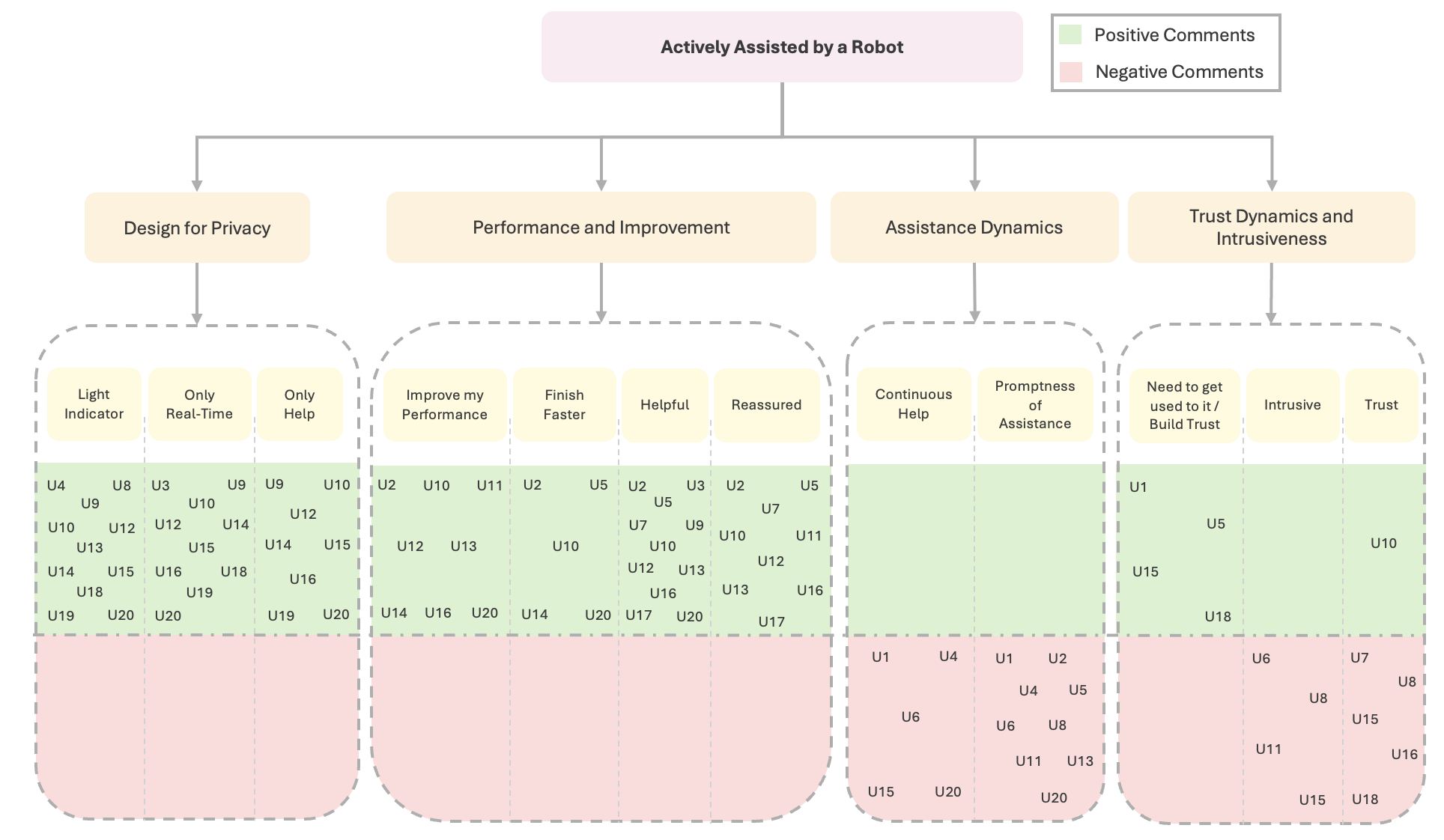}
\caption{The second graph focuses on active robotic assistance and is divided into four branches: (1) \textit{Design of Privacy} (requirements for privacy design), (2) \textit{Performance and Improvement} (effects on user performance and feelings while working with a robot), (3) \textit{Assistance Dynamics} (preferences for help-seeking and assistance), and (4) \textit{Trust Dynamics and Intrusiveness} (trust and comfort with the system).}
\label{fig6}
\end{figure*}

The quotes transcribed into Post-it notes were digitally copied to Excel. This digital transcription enabled further analysis and visualization of the affinity diagram, transformed into two graphs for visual clarity (see Fig.~\ref{fig5}, Fig.~\ref{fig6}).


The affinity diagram revealed that only six participants were open to human assistance, while over half (11 participants) were receptive for robot assistance. Additionally, participants discussed diverse future aspects of robot assistance, leading to three main groups in the diagram: \textit{Assisted on Demand by a Human}, \textit{Assisted on Demand by a Robot} (Fig.~\ref{fig5}), and \textit{Actively Assisted by a Robot} (Fig.~\ref{fig6}). These groups were divided into branches and further into leaves, categorized by positive (green) or negative (red) comments.

Fig.~\ref{fig5} compares perceptions of on-demand assistance from a robot versus a human, divided into three branches: \textit{Emotional Response}, \textit{Communication and Interaction}, and \textit{Assistance Dynamics}.
\textbf{Emotional Response} includes leaves about users' feelings when assisted by a human or the idea of robot assistance. Both groups shared comments on feeling \textit{comfortable}, \textit{fears of bothering}, or \textit{being judged}. For human assistance, there is an additional leaf, \textit{awareness of human feelings}, reflecting how empathy can positively or negatively impact help-seeking. For robot assistance, three unique leaves emerged: \textit{awareness of robots not having feelings}, and feelings of \textit{frustration or annoyance} and \textit{stress} while being assisted by a robot. \textbf{Communication and Interaction} focuses on preferences, communication styles, and issues found during communication. Both groups contains comments regarding \textit{preference} for interaction type and \textit{perceived barriers to asking for help} based on the assistive agent. For human assistance, comments also addressed \textit{communication quality}. For robot assistance, additional leaves included \textit{liking} robot help and the perception that users would have about \textit{asking sooner} and \textit{more frequently} for help, if the assistive agent was a robot. \textbf{Assistance Dynamics} reflects preferences for help-seeking and being assisted. Comments in both groups addressed ideas such as \textit{prompt assistance} and preferences for \textit{being assisted} versus \textit{asking for help}.

Fig.~\ref{fig6} illustrates user perceptions of being actively assisted by a robot, organized into four branches: \textit{Design for Privacy}, \textit{Performance and Improvement}, \textit{Assistance Dynamics}, and \textit{Trust Dynamics and Intrusiveness}. \textbf{Design for Privacy} highlights user concerns about privacy in actively assistive systems. Participants emphasized the need for a \textit{light indicator} to be aware when tracking occurs, stressed the importance of \textit{real-time operation} without recording, and insisted the system should focus solely on \textit{helping} and not performance monitoring. \textbf{Performance and Improvement} reflects user perceptions of how active assistance could enhance task performance. Users found such systems could be \textit{helpful}, enabling them to \textit{improve performance}, \textit{complete tasks faster}, and feel \textit{reassured} by a system adapting to their needs. In \textbf{Assistance Dynamics}, aspects already noted in on-demand assistance groups, are also present here, including concerns about \textit{prompt responses} and apprehensions that \textit{continuous assistance} might limit their ability to work independently. Finally, \textbf{Trust Dynamics and Intrusiveness} captures users trust and comfort levels towards being actively assisted. Some users mentioned needing time to \textit{get used to it}, while others expressed discomfort, finding the system \textit{intrusive}.
%%%%%%%%%%%%%%%%%%%%%%%%%%%%%%%%%%

\section{Discussion}
This study analyzed the challenges users face when seeking help during HRC tasks and evaluated the potential of AI systems to improve these difficulties. Our high-level findings indicate that users frequently struggle to ask for assistance, which is consistent with previous studies \cite{milliken2003exploratory, angelidis2019m}. We further analyze the reasoning behind these difficulties through qualitative data in an HRC scenario. In the near future, robotic assistance may address these issues by using their non-human appearance to reduce social judgment and employing spatial movements to guide users when stuck. However, as detailed in the following subsections, design considerations must ensure cobot assistance is unobtrusive, intuitive, and trusted by users.
%%%%%%%%%%%
\subsection{Reluctance to Ask for Human Help and Potential of Artificial Assistance}
One of the results that aligns with previous research \cite{milliken2003exploratory,angelidis2019m}, is that users are often reluctant to seek help, even when they realize they need it. As noted in Section~\ref{sec4.1}, participants delayed asking for help and in some cases they did not ask for help or asked too late to complete the task successfully. Delays in asking for help sometimes extended over three minutes, during which participants exhibited clear signs of stress and frustration. Through this study, we shed light on the reasons behind participants' hesitation to ask for assistance by analyzing the qualitative data. Twelve participants felt barrier to asking humans for assistance due to fears of bothering others (13 participants) or being judged (7 participants). For example, one noted, \textit{"I don't like asking for help because I feel I disturb the person"} (P01), while another added, \textit{"I feel they might think I am not capable."} (P05). These ideas could be linked to concerns about human sensitivity (7 participants), fearing negative judgments that do not occur with robots. 

In contrast, 17 participants reported no such issues with robot assistance, which was seen as non-judgmental (9 participants) and emotionless (8 participants), qualities that might be attributed to the lack of a human appearance, thereby reducing social judgment, which made users more comfortable seeking help. As participants stated, \textit{"I would prefer to ask for help to a robotic arm, which is like an intelligent tool, than a human that will judge me."} (P13) . These findings suggest that robotic assistance could reduce the reluctance seen with human help, with twenty participants feeling comfortable seeking help from a robot, compared to only eight for humans. Participants also noted that artificial agents would encourage earlier and more frequent help-seeking, as P17 stated: \textit{"I’d ask a cobot for help much sooner and more often because I see it as a reliable tool rather than an assistant who might judge me."}. Additionally, this suggests that effort should not be focused on making robots look human, as their non-human nature already encourages help-seeking behavior.

Based on these insights, several participants offered design suggestions to enhance AI assistance in HRC scenarios. In such scenarios, users pointed out the need for a clear signal indicating that the robot is open to assist, similar to how humans use eye contact. One participant remarked, \textit{"I would ask also sooner if I see that the robot is open to help me, as a person might make eye contact with me to demonstrate it."} (P11). In the same line, five users recommended a color change to indicate it is already in assistive mode, while continuing its work. Three participants noted that this feature would help prevent the feeling of “bothering” the robot while working with it, as captured by P17: \textit{"I would like that once I ask for help, the robot starts talking to me, but while it keeps working. If the robot stops I would feel that I am disturbing it."}.

However, not all aspects of robotic assistance were viewed positively. In the On-Demand Assistance by a robot scenario, four participants reported they could feel frustrated, and more stressed knowing a robot would assist them. For example, P08 remarked, \textit{"Knowing that a robot would assist me would stress me, and it would be very annoying if it doesn’t provide the help I need."}.
%%%%%%%%%%%%%%%%%%%
\subsection{Impact on Task Performance and User Experience}
Hesitation to seek help negatively impacted both task performance and user experience. This resulted in incorrect or incomplete assemblies and wasted time as users attempted to recover without requesting help (see Section~\ref{sec4.1} for quantitative data), with two users failing to complete the task and two others who completed it but without the missing piece. Two of those four even stared at the robot looking frustrated.

The prolonged stress created by these situations, along with the hesitation and need for help seen with human assistance, could be mitigated by artificial agents, as participants expressed a strong interest in proactive robotic support compared to human assistance. With human assistance, most participants (14 out of 20) preferred asking for help rather than being assisted (6 out of 20), despite the difficulties in requesting help. In contrast, when the robot is the assistant, the same number of participants (11 participants) liked the idea of asking for help and being actively assisted. Fig.~\ref{fig6} supports this trend, with users believing that in HRC tasks, with proactive robotic assistance, they would have improved their performance (8 participants), finished faster (5 participants), felt reassured (9 participants), and generally found the robot assistance helpful (11 participants). However, users remarked that a key point in the proactive assistance is in the way that robots provide help. Seven participants preferred to receive hints over complete help, with five participants pointing out that they preferred robots to use their spatial capability to point out the desired piece (\textit{"I would like that the robot instead of bringing me a piece, points me out where is the piece."} (P09)). The other two expressed they would just prefer general hints such as \textit{"I would like that the robot just tells me I am going in the wrong direction."} (P20).

Additionally, users suggested proactive assistance in collaborative scenarios should begin with a brief conversation to assess the worker’s state (2 participants), to identify users needs and determine the appropriate level of assistance required on a daily basis (5 participants) (\textit{"Everyday is different and I would like that the robot could learn each day how to assist me."} (P05)). Additionally, two participants noted that for the initial interaction, they preferred the robot to stay stationary and simply converse rather than moving, in contrast to periods of active collaboration. They attributed this to the robot's limited human-like qualities, noting that a moving robot might seem impersonal or "cold" on first contact. P10 explained,\textit{ "For a first contact, I would like it to be stopped and have a light indicating it is talking to me. If it were working, that would be cold."}. Furthermore, some users noted that in collaborative tasks for the robot to serve as a reliable companion, it should learn and recognize each worker's unique working habits (2 participants).

In contrast to the positive comments, some participants expressed concerns about being actively assisted by the robot, noting the technology could feel intrusive or evoke mistrust. Four participants mentioned needing time to adjust, believing that human assistance offers superior quality of communication due to empathy. As P12 explained, \textit{"I feel more comfortable with a human since I don’t know how to express my needs to an artificial agent."}. Additionally, some participants worried about the dynamics of assistance when collaborating, with five expressing apprehension about continuous help and nine about overly prompt responses. As remarked by six users, prompt assistance can be annoying and make the collaboration less efficient, and when a worker is an expert, they might prefer that the robot waits for longer time before offering help (\textit{"A worker with expertise will not like that the robot is helping him very soon."} P20).

To build trust in robotic assistance, it is crucial to address privacy design concerns. Participants recommended including a light indicator to show when the system is active, ensuring it operates only in real-time without recording, and emphasizing that it is used solely to provide help rather than track performance. As P10 stated, \textit{"I would like it for helping me, but not for tracking my performance."}.
%%%%%%%%%%%%%%%%%%
\section{Design Guidelines for an Assistive Robot} 
Based on our findings, key design guidelines emerge for developing an initial assistive robot for HRC tasks:

\textit{Proactive and Non-Intrusive Assistance:} The robot should detect when a user needs help and offer assistance in a non-intrusive manner. An indicator light or sound could notify users that the robot is aware they might need help, allowing them to continue working without feeling disrupted. Such signaling may also encourage users to seek help sooner. The indicator should remain active while the robot is assisting, ensuring that the robot continues working as it provides help.

\textit{Neutral, Non-Judgmental Support:} In HRC scenarios, where asking for help can feel disruptive, a neutral, consistent tone can reassure users and encourage them to interrupt the workflow. Keeping the robot active during assistance can reinforce this effect further.

\textit{Customizable Assistance Levels:} Since individual assistance preferences vary, the robot should let users customize help levels for a more human-centric interaction, while intervening if prolonged stress or frustration is detected.

\textit{Encouraging Autonomy and Learning:} In HRC tasks, support should nurture autonomy without disrupting workflow. Offering hints or partial solutions can make users solve problems independently, while spatial cues (pointing out a piece), offer subtle guidance. Gradually reducing assistance can build confidence, enhance learning, and minimize fear of failure.

\textit{Privacy and Transparency:} Users emphasized the need for a light indicator to signal when the system is active and assurances that the system operates in real-time without recording. It should solely assist workers and not track performance metrics. Companies should provide clear documentation committing to these privacy principles.

These guidelines outline key features for designing an assistive robot that meets user needs and preferences, enhancing both user experience and performance in HRC scenarios.

%%%%%%%%%%%%%%%%%%%%%%%%%%%%%%%%%

\section{Limitations and Future Work}
While this study offers valuable insights into human assistance in HRC tasks and user attitudes toward robotic assistance, it is limited to human assistance. Future research should apply these findings to develop robotic assistance for HRC tasks and examine the impact on user behavior, performance, and related metrics. Exploring different types of assistance, such as active versus on-demand, could provide further insights into how each aspect affects performance and user experience. A user study comparing human on-demand assistance, robot on-demand assistance, and active robot assistance would deepen understanding of user needs and improve interactions.

Additionally, the study took place in a controlled quiet laboratory environment, using an assembly task representative of industrial work. Expanding this research to real industrial settings with diverse assembly tasks could provide broader insights into user behavior when seeking help.

Finally, future work should focus on developing algorithms to automatically detect when a user needs help. 

%%%%%%%%%%%%%%%%%%%%%%%%%%%%%%%%

\section{Conclusions}

Our study reveals that users, who often hesitate to seek human help due to fears of judgment and disruption, prefer proactive, non-judgmental robotic assistance to enhance performance in HRC tasks. Proactive robotic assistance was particularly well-received, with users indicating it could enhance performance, speed up task completion, and provide greater reassurance, underscoring its potential to improve user experience in HRC tasks. Users also noted that while robots should exhibit humanized cues when offering and providing help, they must still retain their operational efficiency. However, the study also revealed concerns about robotic assistance, including intrusiveness, mistrust, and the need for adaptation to new technology. To address these challenges, future designs should focus on enhancing communication quality, ensuring robots provide support in a natural and timely manner while avoiding continuous or overly immediate responses and take into account privacy considerations. Our findings yield preliminary guidelines for designing cobots that effectively assist humans in HRC tasks.

\addtolength{\textheight}{-12cm}   % This command serves to balance the column lengths

%%%%%%%%%%%%%%%%%%%%%%%%%%%%%%%%%%%%%%%%%%%%%%%%%%%%%%%%%%%%%%%%%%%%%%%%%%%%%%%%

%%%%%%%%%%%%%%%%%%%%%%%%%%%%%%%%%%%%%%%%%%%%%%%%%%%%%%%%%%%%%%%%%%%%%%%%%%%%%%%%

%%%%%%%%%%%%%%%%%%%%%%%%%%%%%%%%%%%%%%%%%%%%%%%%%%%%%%%%%%%%%%%%%%%%%%%%%%%%%%%%

%%%%%%%%%%%%%%%%%%%%%%%%%%%%%%%%%%%%%%%%%%%%%%%%%%%%%%%%%%%%%%%%%%%%%%%%%%%%%%%%
\bibliographystyle{IEEEtran}  
\bibliography{bibliography.bib}

\end{document}